\documentclass{article}
\PassOptionsToPackage{numbers, compress}{natbib}

\usepackage[preprint]{neurips_2019}

\usepackage[utf8]{inputenc} 
\usepackage[T1]{fontenc}    
\usepackage{hyperref}       
\usepackage{url}            
\usepackage{booktabs}       
\usepackage{amsfonts}       
\usepackage{nicefrac}       
\usepackage{microtype}      
\usepackage{graphicx}
\usepackage{amsmath}
\usepackage{multirow}

\title{Salient Object Detection Combining a Self-attention Module and a Feature Pyramid Network}

\author{%
  Guangyu Ren\\
  Department of Electrical and Electronic Engineering\\
  Imperial College London\\
  \texttt{g.ren19@imperial.ac.uk} \\
  \And
  Tianhong Dai\\
  Department of Bioengineering\\
  Imperial College London\\
  \texttt{tianhong.dai15@imperial.ac.uk} \\
  \AND
  Panagiotis Barmpoutis \\
  Department of Electrical and Electronic Engineering \\
  Imperial College London \\
  \texttt{p.barmpoutis@imperial.ac.uk} \\
  \And
  Tania Stathaki \\
  Department of Electrical and Electronic Engineering \\
  Imperial College London \\
  \texttt{t.stathaki@imperial.ac.uk} \\
}

\begin{document}

\maketitle

\begin{abstract}
  Salient object detection has achieved great improvement by using the Fully Convolution Network (FCN). However, the FCN-based U-shape architecture may cause the dilution problem in the high-level semantic information during the up-sample operations in the top-down pathway. Thus, it can weaken the ability of salient object localization and produce degraded boundaries. To this end, in order to overcome this limitation, we propose a novel pyramid self-attention module (PSAM) and the adoption of an independent feature-complementing strategy. In PSAM, self-attention layers are equipped after multi-scale pyramid features to capture richer high-level features and bring larger receptive fields to the model. In addition, a channel-wise attention module is also employed to reduce the redundant features of the FPN and provide refined results. Experimental analysis shows that the proposed PSAM effectively contributes to the whole model so that it outperforms state-of-the-art results over five challenging datasets. Finally, quantitative results show that PSAM generates clear and integral salient maps which can provide further help to other computer vision tasks, such as object detection and semantic segmentation.
\end{abstract}

\section{Introduction}
Salient object detection or segmentation aims to identify visually distinctive parts of a natural scene. With this capability of providing high-level information, the saliency detection is widely applied in the computer vision applications, such as object detection~\cite{ren2013region, zhang2017bridging, liu2019gated} and tracking~\cite{hong2015online}, visual robotic manipulations~\cite{yuan2018rgb, schillaci2013evaluating}, image segmentation~\cite{wei2017object, wang2018weakly} and video summarization~\cite{ma2002user, simakov2008summarizing}. In early studies, the salient object detection was formulated as a binary segmentation problem. However, the connections' establishment between the salient object detection and other computer vision tasks was unclear. Nowadays, convolution neural network (CNN) attracts more attention in the research community. Compared with the classic hand-crafted feature descriptors~\cite{lowe2004distinctive, dalal2005histograms}, CNNs have stronger feature representation ability. Specifically, CNN kernel with small receptive fields can provide local information and the kernel with large receptive fields can provide global information. This characteristic enables CNN-based approaches to detect salient areas with refined boundaries~\cite{borji2014salient}. Thus, CNN-based approaches have become the major research field in the salient object detection.

Recently, Fully Convolution Networks (FCNs) becomes the fundamental framework in the salient object detection~\cite{liu2019simple, wang2019salient, qin2019basnet}, as FCNs can be fed by arbitrary size of input and achieve richer spatial information compared with the fully connected layer. Although these works have achieved great improvement in the performance, they are still restricted by some limitations. FCN-based approaches utilize multiple convolution layers and pooling layers to produce the high-level semantic features which are helpful to locate objects but they may lose information during pooling operations. This can lead to degraded boundaries of detected objects being generated. Besides, when the high-level features are upsampled to generate score prediction for each pixel, it will also be diluted which could decrease the ability of object localization. 

\begin{figure}[h]
\minipage{0.19\textwidth}
  \centering
  \includegraphics[width=\linewidth]{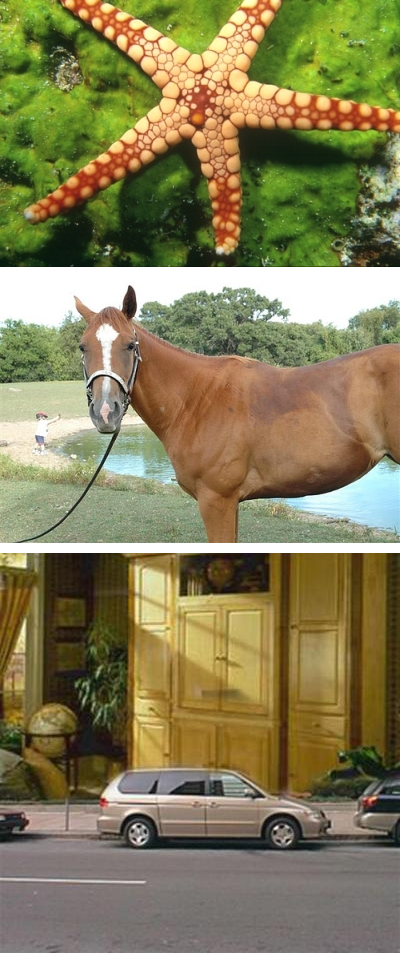}
  (a) Image
\endminipage\hfill
\minipage{0.19\textwidth}
  \centering
  \includegraphics[width=\linewidth]{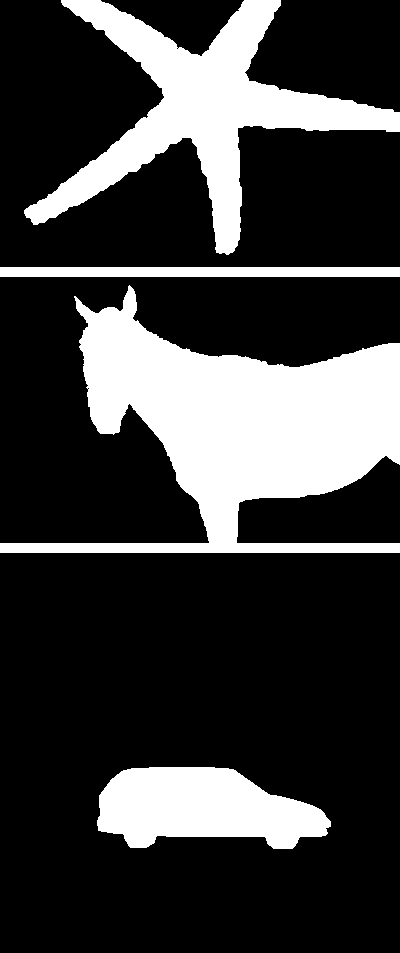}
  (b) ground truth
\endminipage\hfill
\minipage{0.19\textwidth}%
  \centering
  \includegraphics[width=\linewidth]{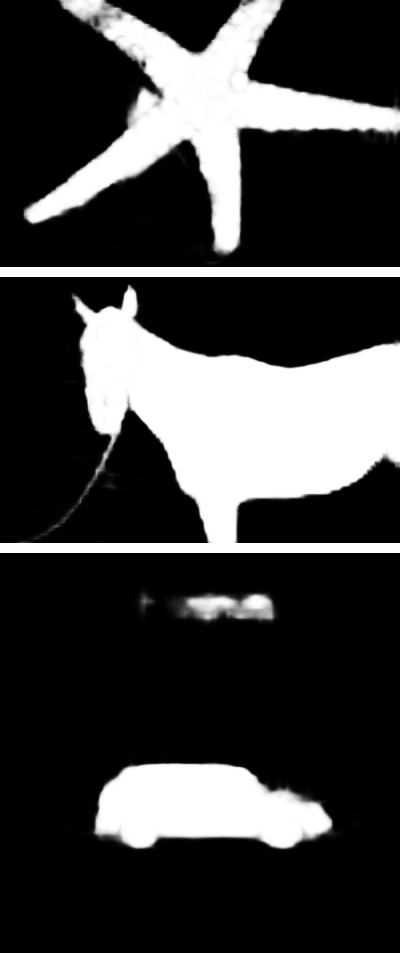}
  (c) baseline
\endminipage\hfill
\minipage{0.19\textwidth}%
  \centering
  \includegraphics[width=\linewidth]{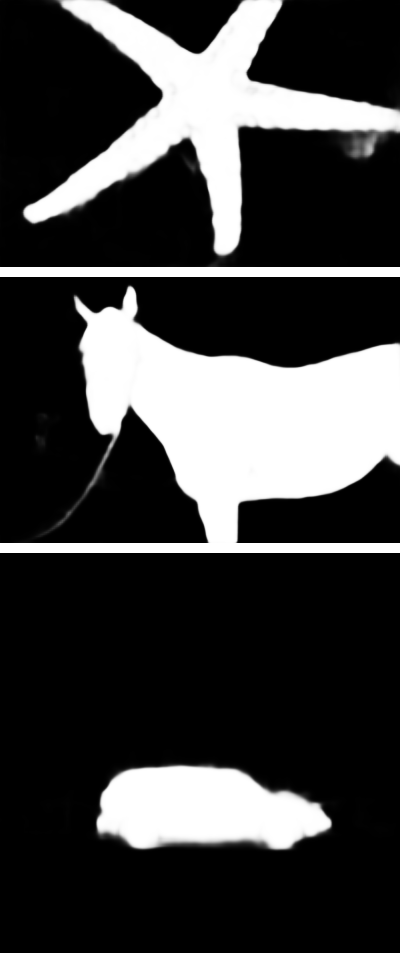}
  (d) baseline+sa
\endminipage\hfill
\minipage{0.19\textwidth}%
  \centering
  \includegraphics[width=\linewidth]{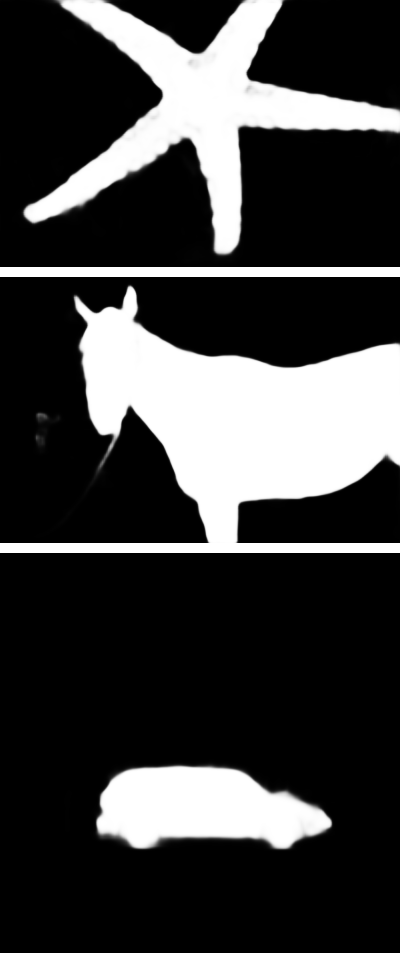}
  (e) ours
\endminipage\hfill
\caption{Qualitative visual results of ablation studies. (a) original images, (b) ground truth, (c) baseline, (d) baseline+self-attention (SA), (e) our method.}
\label{fig:ablation}
\end{figure}

In this paper, we propose a novel pyramid self-attention module (PSAM) to overcome the limitation of feature dilution of the previous FCN-based approaches. Figure~\ref{fig:ablation}(c) shows the inherent problems of Feature Pyramid Networks (FPNs). Through incorporating self-attention module with multi-scale feature map of FPNs, the model will focus on the high-level features. This leads to the extraction of features with richer high-level semantic information and larger receptive fields. In addition, a channel-wise attention module is employed to reduce the redundancy in the FPN which can refine the final results. Experimental results show that PSAM can improve the performance of salient object detection and achieve state-of-the-art results in five challenging datasets. The contributions of this work can be concluded as: 1) We propose a novel pyramid self-attention structure which can make the model focus more on the high-level features and reduce feature dilation in top-down pathway. 2) We adopt a channel-wise attention to reduce the redundant information in the lateral connections of the FPN to refine the final results.
\section{Related Works}
\paragraph{Salient Object Detection} Due to the outstanding feature representation ability of CNN, the hand-craft feature based methods have been replaced by the CNN models. In the work of ~\cite{li2015visual}, Li and Yu used fully connected layers on the top CNN layers to extract different scale features of a single region. Then, multi-scale features were used to predict the scores for each region. ~\cite{zhao2015saliency} utilized two independent CNN to extract the global context of the full image and the local context of the detailed area to train the model jointly. However, the spatial information was lost in these CNN-based methods, because of the fully connected layers.

Recently, FCN-based methods have raised more concerns in the salient object detection. ~\cite{qin2019basnet} proposed a boundary-aware salient object detection network which incorporates a predict module and a residual refinement module module (RRM). The predict module was used to estimate the salient map from the raw images and the RRM was used to refine the results from the predict module which was trained by using the residual between the salient map and the ground truth. ~\cite{liu2019simple} introduced a PoolNet structure which has two pooling modules: global guidance module (GGM) and feature aggregation module (FAM). The GGM was designed to acquire more high-level information around the inputs which tackles the feature dilation problem in the U-shape network structure. Then, the FAM merges the multi-scale features of the FPN, leading to reduce the problem of aliasing caused by the up-sampling and enlarge the receptive fields. From the experiments, PoolNet can make more precise localization of the sharpened salient objects compared with other baseline approaches. In the work of ~\cite{wu2019cascaded}, a Cascaded Partial Decoder (CPD) structure that contains two prime branches was proposed. The first branch contributes in the computation speed improvement by dropping features in the shallow layers. The second branch uses the salient map from the first branch in order to refine the features in the deeper layers which ensures the speed and accuracy of the framework.

\paragraph{Attention Mechanism} Attention mechanism is mainly used in the area of Natural Language Processing (NLP). ~\cite{vaswani2017attention} introduced a framework called Transformer which was used to replace the recurrent layers through using attention mechanism to capture global dependencies between input and output. This framework also allows parallel computing which leads to faster speed compared to recurrent networks. Except for the sequence models, this kind of attention mechanism is also needed in the CNN models. Different from the attention mechanism in the sequence models, self-attention was introduced to utilize attention mechanism in the single context data. ~\cite{bello2019attention} proposed Attention Augmented Convolutional Network which produces attentional feature maps via self-attention module and combines these with CNN feature maps to capture spatial dependencies of the input and it achieves huge improvement in the tasks of object classification and detection. The stand alone self-attention layer was introduced in the work of ~\cite{ramachandran2019stand}. It can be used to set up a fully attention model through replacing all spatial convolution layers with self-attention layers. The self-attention layer leverages the components in the previous works and proves that it can be used as a stand-alone layer which can replace the spatial convolution layer easily. 
\begin{figure}[h]
\centering
\includegraphics[width=0.9\columnwidth]{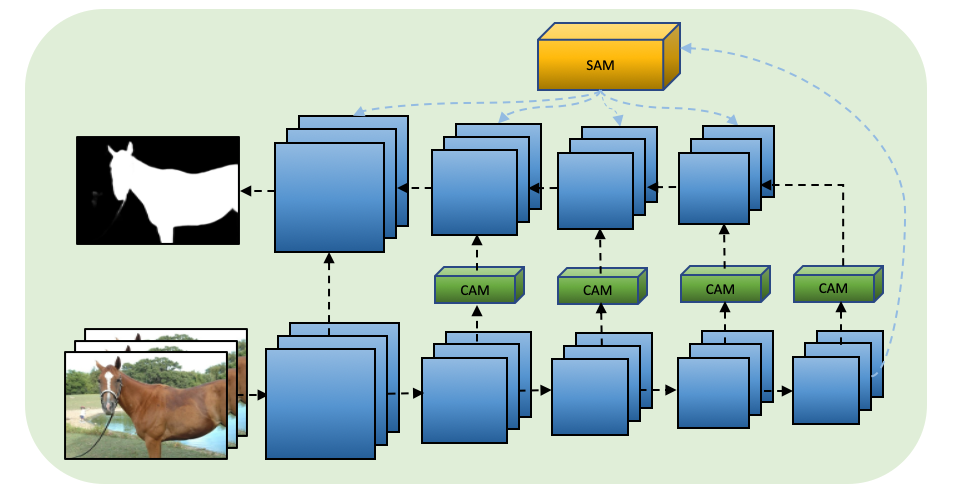}
\caption{Overall pipeline of the proposed model}
\label{fig:overall}
\end{figure}
\section{Self-Attention Based FPN}
In this section, we describe the proposed architecture that integrates two attention modules. More specifically, we use a pyramid self-attention module which aims to enhance the high-level semantic features and transmit the enhanced semantic information to different feature levels. In addition, when feature maps are merged in the top-down pathway, a simple channel-wise attention~\cite{hu2018squeeze, chen2017sca, zhao2019pyramid} module is added in each lateral connection to focus on the high responses of salient objects. 
The proposed architecture is based on a classic feature pyramid network (FPNs)~\cite{lin2017feature} which exploits ResNet~\cite{he2016deep} as a backbone. It is well known that this basic FPN architecture has been widely used in many different computer vision tasks, especially for detection tasks, leading to accurate detection results because of its robust and reasonable structure. As shown in Figure~\ref{fig:overall}, we retain the basic structure and introduce two effective modules to achieve a state-of-the-art performance. A pyramid self-attention module, which is built between the bottom-up and top-down pathway, supports the model to focus on the high-level features which contain semantic information. Then this module transfers the processed high-level information to each feature levels in the top-down path. Meanwhile, feature maps from different stages in ResNet pass through a channel-wise attention module to further emphasize the context information.

\begin{figure}[h]
\centering
\includegraphics[width=0.85\columnwidth]{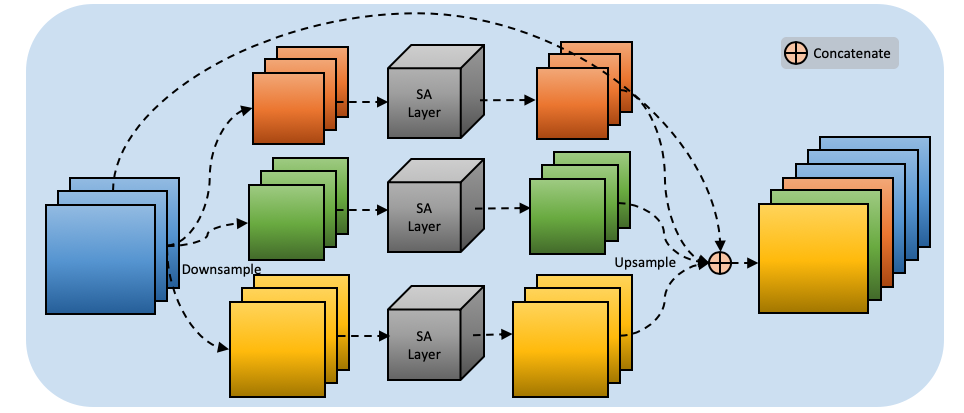}
\caption{The structure of Pyramid Self-attention Module.}
\label{fig:PSAM}
\end{figure}
\label{sec:self-attention}
\subsection{Pyramid Self-Attention Module}
In this subsection, we describe the proposed module in detail and demonstrate the differences from previous works. ~\cite{hou2017deeply} has demonstrated that high-level semantic features are more representative and discriminative, leading to the position of salient objects being located more accurately. Figure~\ref{fig:ablation}(c) shows that without any extra attention modules, the FPN baseline can generate rough saliency map which has insufficient and incomplete salient objects. Meanwhile, there are also some non-salient objects which should not be detected in the saliency maps. These error predictions are caused by two main challenges which cannot be avoided in the FPN architecture. The first problem is that the high-level information is diluted progressively when it is integrated in different feature levels in the top-down pathway. Another problem of the FPN baseline is that this architecture can be impacted by other non-essential information which may reduce the final performance of the model. In other words, the FPN architecture detects not only incomplete salient objects but also unnecessary objects. To overcome these two intrinsic problems of the baseline, we propose a novel pyramid self-attention module (PSAM) which contains stand-alone self-attention layers~\cite{ramachandran2019stand} in different scales, further focusing on important regions and enlarging the receptive field of the model.
Specifically, as shown in Figure~\ref{fig:PSAM}, PSAM firstly transforms the feature map which is produced by bottom-up pathway into multi-scale feature regions and then each self-attention layer learns to pay more attention on important semantic information. After processed by self-attention layers, these multi-scale representations, which contain effective semantic information, are concatenated together to complement high-level semantic information in the top-down pathway. More technically, let $X^{in}$ denote the feature map, which is produced by the top-most layer. We downsample the feature map $X^{in}\in R^{H\times\ W\times C}$ into three different scales denoted as $\{X_{1}, X_{2}, X_{3}\}$. Given a pixel $x_{ij}\in\{X_{1},X_{2},X_{3}\}$ a corresponding local memory block $r_{k}$ is extracted from the same feature map $X_{i}\in\{X_{1}, X_{2}, X_{3}\}$. This $r_{k}$ is a $k\times k$ region which surrounds $x_{ij}$. There are three crucial learnable parameters in this self-attention algorithm: queries, keys and values. We use $W_{Q}$, $W_{K}$ and $W_{V}$ to represent their learnable weights respectively. The final attention output pixel is computed as follows:
\begin{equation}
    \centering
    s_{ij} = \sum_{i,j\in N_{k}}\text{softmax}(q_{ij}^{T}k_{r_{k}})v_{r_{k}},
\end{equation}
where $q_{ij}=W_{Q}x_{ij}$, $k_{r_{k}}=W_{K}x_{r_{k}}$, $q_{r_{k}}=W_{V}x_{r_{k}}$ denote the three crucial parameters, $s_{ij}$ denotes the output pixel of a self-attention layer, $N_{k}=\{(i, j): i=\{-\frac{k-1}{2}, ..., \frac{k-1}{2}\},$ $j=\{-\frac{k-1}{2}, ..., \frac{k-1}{2}\}\}$ defines the coordinates of $r_{k}$ and we use $\{Y_{1},Y_{1},Y_{3}\}$ to denote the final feature map which further represents an upsampling operation after each self-attention layer. Then we concatenate them with the original $X^{in}$ to generate the final output of the PSAM.
\begin{equation}
    \centering
    Y^{out} = \text{concat}\{Y_{1},Y_{1},Y_{3}, X^{in}\}
\end{equation}
Inspired by the previous work~\cite{liu2019simple}, we exploit a similar complementing strategy to avoid the dilution of high-level semantic information. However, compared to the previous work, the proposed pyramid self-attention module achieves a state-of-the-art performance. This module is built at the end of bottom-up pathway and converts the high-level semantic features into different scales, further enlarging the receptive field of the model. Based on multi-scale high-level feature maps, the attention layers view these semantic features at different scales and then can achieve a comprehensive attention task. More performance detail will be shown in the ablation study.
\begin{figure}[h!]
\centering
\includegraphics[width=0.85\columnwidth]{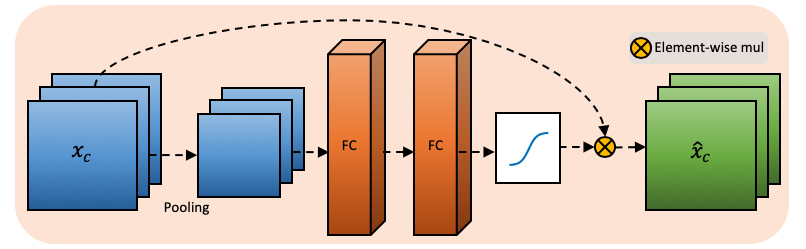}
\caption{The structure of Channel-wise attention Module.}
\label{fig:channel}
\end{figure}
\subsection{Channel-Wise Attention}
To enhance the context and structural information, the lateral connection has been used in the top-down pathway, leading to a state-of-the-art performance of detection tasks. However, this operation also introduces some unmeaningful information, which can reduce the performance and impact on the final prediction. From Figure~\ref{fig:ablation}(c), the two problems caused by features redundancy are obvious. The first problem is that there are extra regions which should not be detected in the saliency map. Another problem is that the edges of salient objects are ambiguous. Both problems indicate that a further refinement should be applied. 
~\cite{hu2018squeeze} has pointed out that different channels have different semantic features and channel-wise attention can capture channel-wise dependencies. In other words, channel-wise attention can emphasize the salient objects and alleviate the inaccuracy which is caused by redundant features in channels. Therefore, we add this simple channel-wise attention~\cite{hu2018squeeze, chen2017sca, zhao2019pyramid} to each later connection to achieve a refinement task. 
The structure of channel-wise attention is shown in Figure~\ref{fig:channel}. It consists of one pooling layer and two fully-connected layers which are followed by a ReLU~\cite{nair2010rectified} and a sigmoid function respectively. First, an operation of squeezing global spatial information is applied to each channel. This step can be easily implemented by an average pooling : 
\begin{equation}
    \centering
    p_{c} = F_{pool}(X_{c}) = \dfrac {1}{H\times W}\sum ^{H}_{i=1}\sum ^{W}_{j=1}x_{c}\left( i,j\right)
\end{equation}
where c refers to the channel number, H x W refers to the spatial dimensions of i-th element of $X_{c}$. 
After the pooling operation, the generated channel descriptor is fed into the fully-connected layers to fully capture channel-wise dependencies. 
\begin{equation}
    \centering
    s_{c} = F_{fc}(p_{c},W) = \sigma(fc_{2}(\delta(fc_{1}(p_{c},W_{1})),W_{2})
\end{equation}
where $\sigma$ refers to the sigmoid function, $\delta$ refers to the ReLU function and fc means the fully-connected layers. Finally, this generated scalar $s_{c}$ multiplies the feature map $X_{c}$ to generate a weighted feature map $\hat{X_{c}}$:
\begin{equation}
    \centering
    \hat{X_{c}} = s_{c}\cdot X_{c}
\end{equation}
\section{Experiments}
\subsection{Datasets and Evaluation Metrics}
For the evaluation of the proposed methodology, we carry out a series of experiments using five popular saliency detection benchmarks. More specifically, we use the: ECSSD~\cite{yan2013hierarchical}, DUT-OMRON~\cite{yang2013saliency}, DUT-TE~\cite{wang2017learning}, HKU-IS~\cite{li2015visual} and SOD~\cite{movahedi2010design}. These five datasets consist of a variety of objects and structures which are still challenging for salient object detection algorithms to locate and detect them precisely. For the training of our model we use the large-scale dataset DUTS~\cite{wang2017learning}, which contains 10533   training images and 5019 testing images.
To evaluate the performance of the model, we estimate three representative evaluation metrics: precision-recall curves, F-measure score and mean absolute error (MAE). F-measure indicates the standard overall performance which are computed by precision and recall:
\begin{equation}
    \centering
    F_{\beta }=\dfrac {\left( 1+\beta ^{2}\right) \cdot \text{precision}\cdot \text{recall}}{\beta ^{2}\cdot \text{precision}+\text{recall}}
\end{equation}
where $\beta^{2}$ is set to 0.3 as default, precision and recall are obtained by using different thresholds to compare prediction and ground truth.
The MAE indicates the deviations between the binary saliency map and the ground truth. In other words, this metric quantifies the similarity between prediction map and ground truth mask:
\begin{equation}
    \centering
    MAE = \dfrac {1}{W\times H}\sum ^{W}_{x=1}\sum ^{H}_{y=1}\left| P\left( x,y\right) -G\left( x,y\right) \right|
\end{equation}
where W denotes the width and H denotes the height of prediction, P denotes the prediction map which is the output of the model and G represents the ground truth.

\begin{figure}[h]
\minipage{0.32\textwidth}
  \centering
  \includegraphics[width=\linewidth]{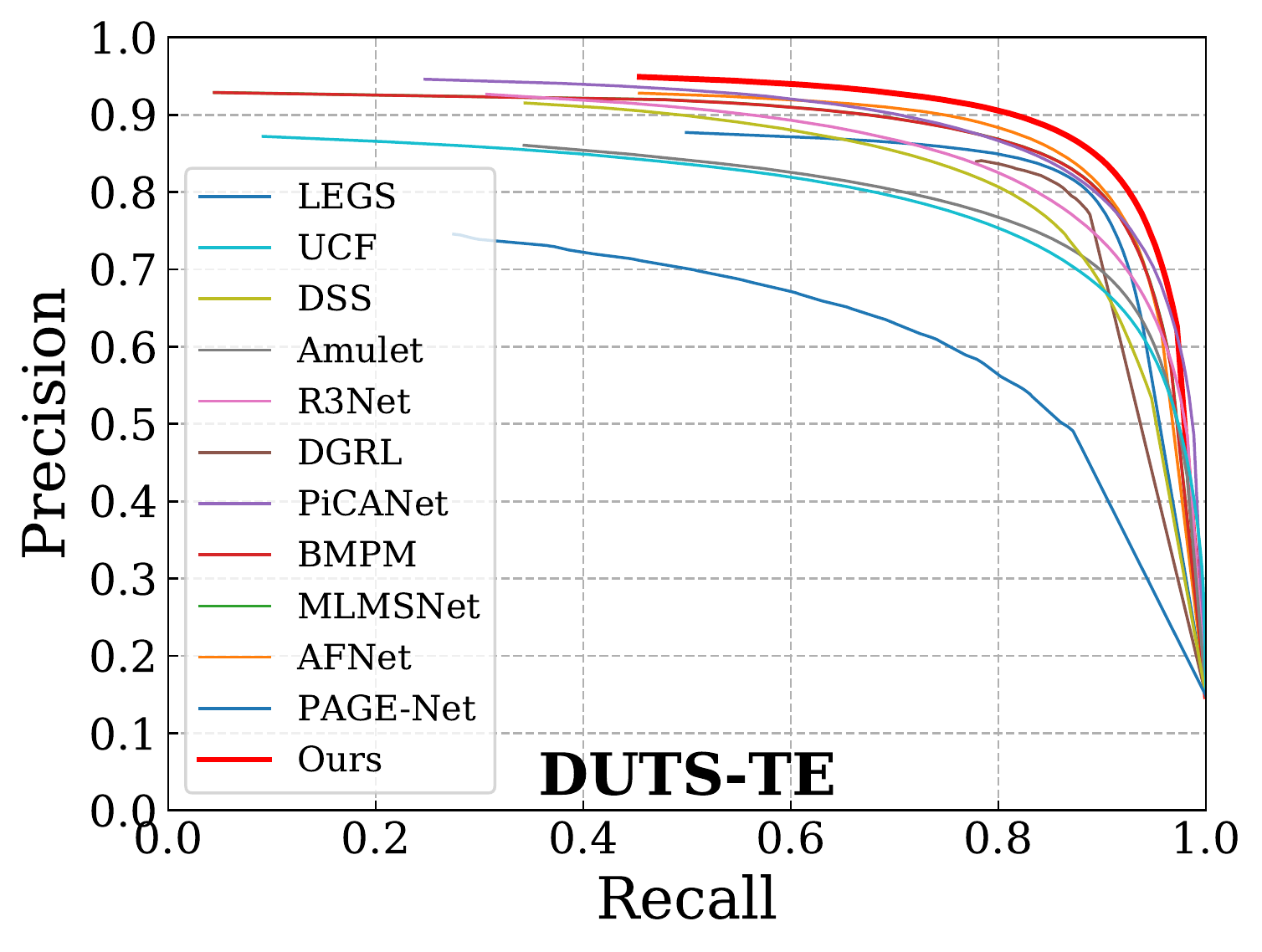}
  (a) DUTS-TE
\endminipage\hfill
\minipage{0.32\textwidth}%
  \centering
  \includegraphics[width=\linewidth]{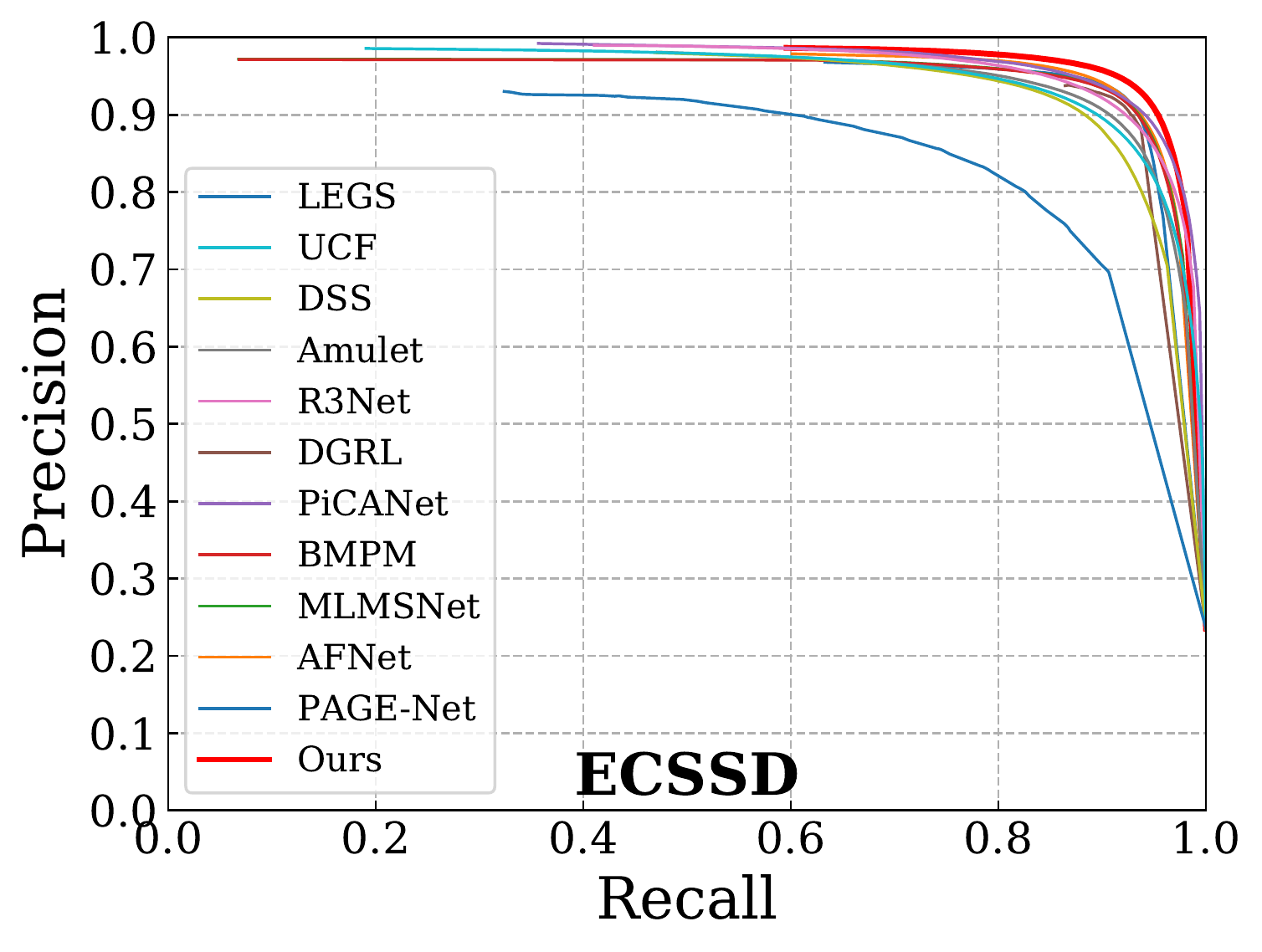}
  (b) ECSSD
\endminipage\hfill
\minipage{0.32\textwidth}%
  \centering
  \includegraphics[width=\linewidth]{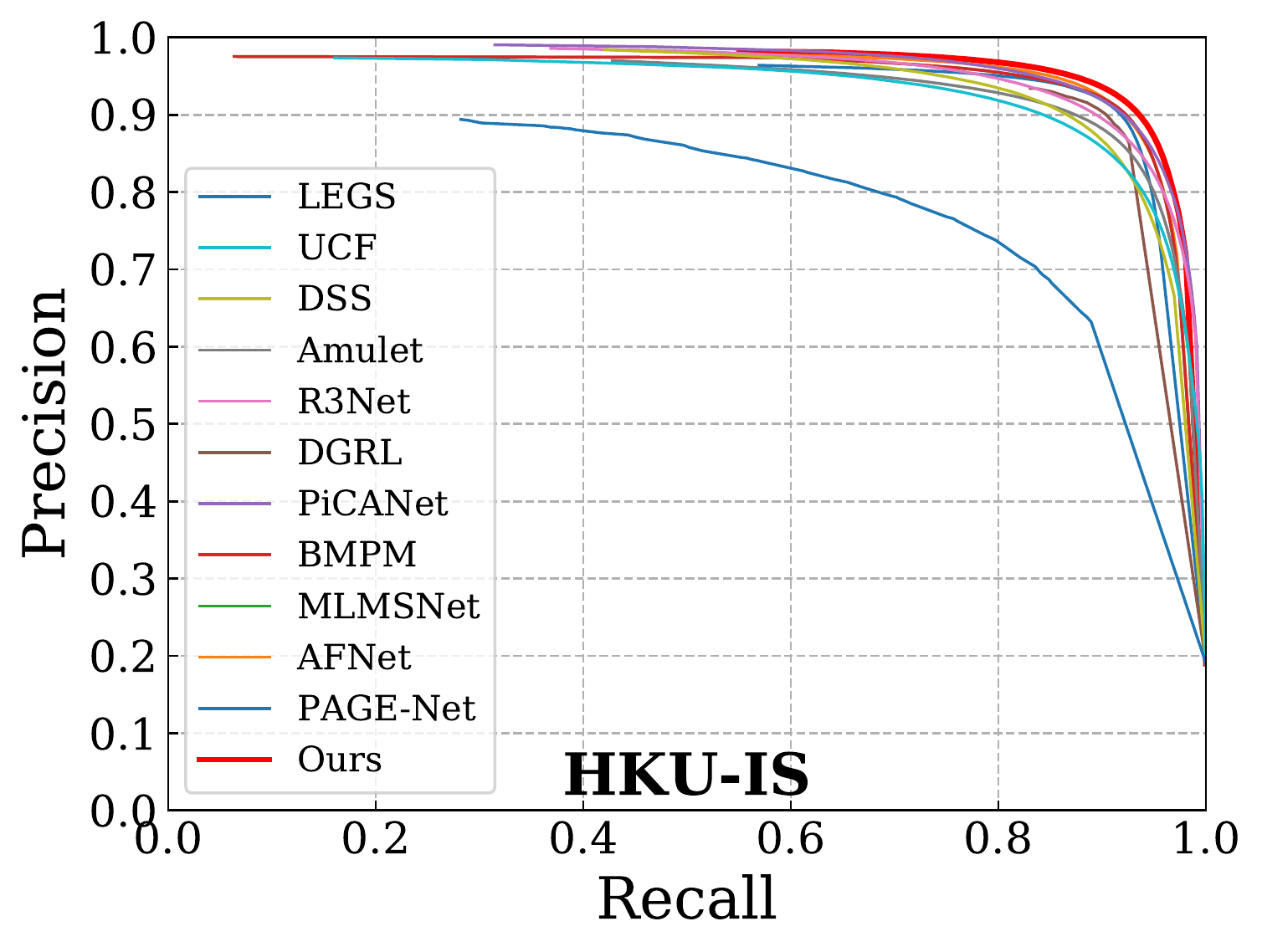}
  (c) HKU-IS
\endminipage\hfill
\minipage{0.32\textwidth}%
  \centering
  \includegraphics[width=\linewidth]{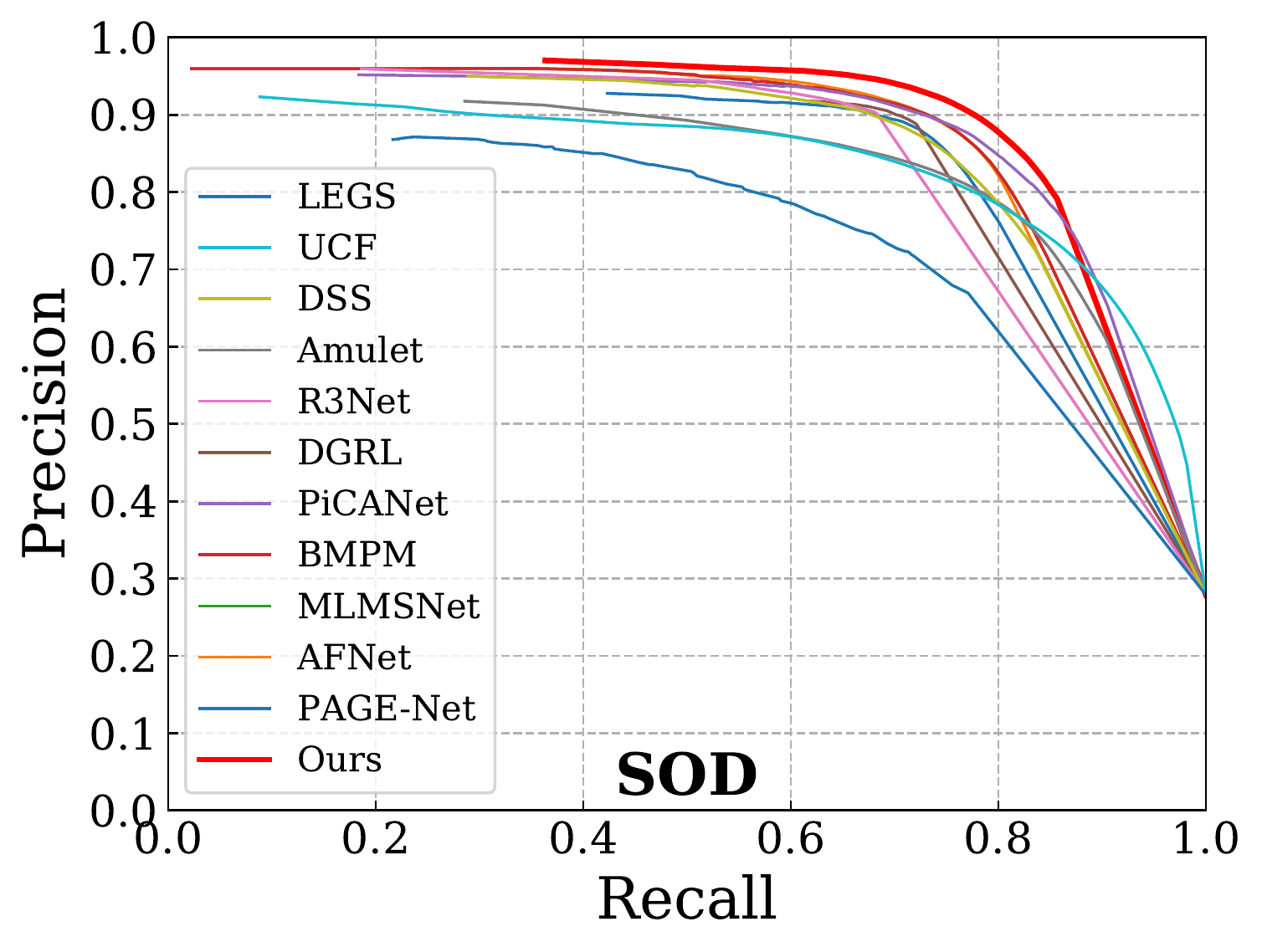}
  (d) SOD
\endminipage\hfill
\minipage{0.32\textwidth}%
  \centering
  \includegraphics[width=\linewidth]{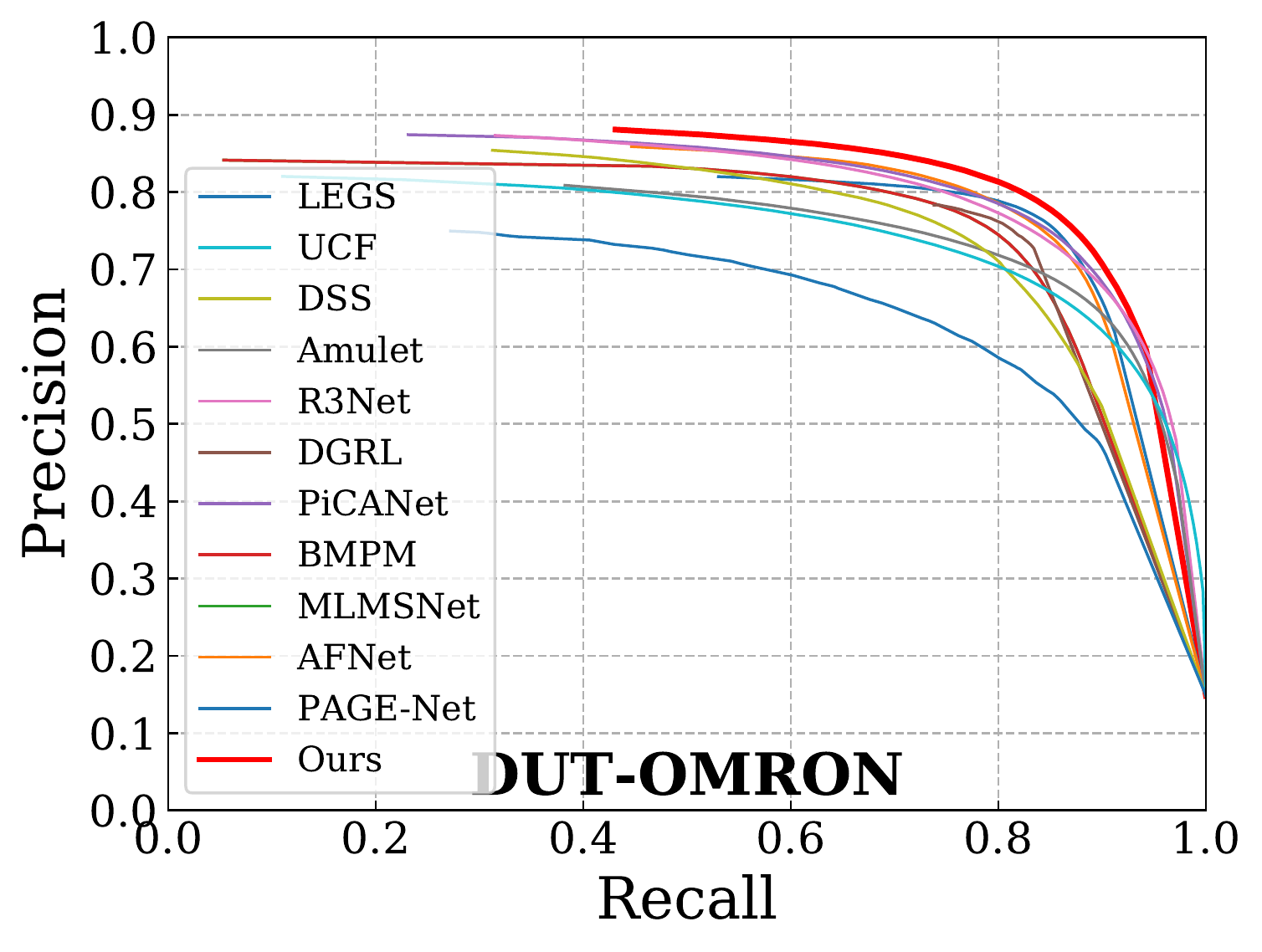}
  (e) DUT-OMRON
\endminipage\hfill
\minipage{0.32\textwidth}%
  \centering
  \includegraphics[width=\linewidth]{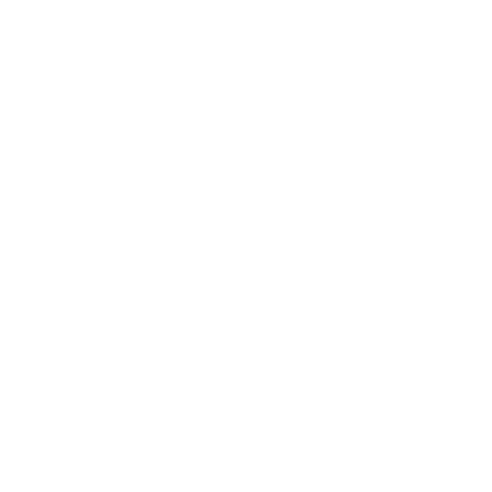}
\endminipage\hfill
\caption{Results with PR curves on five benchmark datasets: DUTS-TE, ECSSD, HKU-IS, SOD and DUT-OMRON. x-axis represents the recall rate and y-axis represents the precision.}
\label{fig:pr_curves}
\end{figure}
\subsection{Impelmentation Detail}
Our model is implemented in Pytorch. We use ResNet-50~\cite{he2016deep} as a backbone which has been pre-trained on ImageNet~\cite{krizhevsky2012imagenet}. The proposed architecture is trained on a GTX TITAN X GPU for 24 epochs. As suggested in~\cite{liu2019simple}, the initial learning rate is set equal to 5e-5 for the first 15 epochs and then reduces to 5e-6 for the last 9 epochs. We adopt 0.0005 weight decay for the Adam~\cite{kingma2014adam} optimizer and binary cross entropy loss function in the proposed framework. Finally, in order to increase the robustness of the model, we perform data augmentation through the application of random horizontal flipping.
\begin{figure}[h]
\centering
\includegraphics[width=\columnwidth]{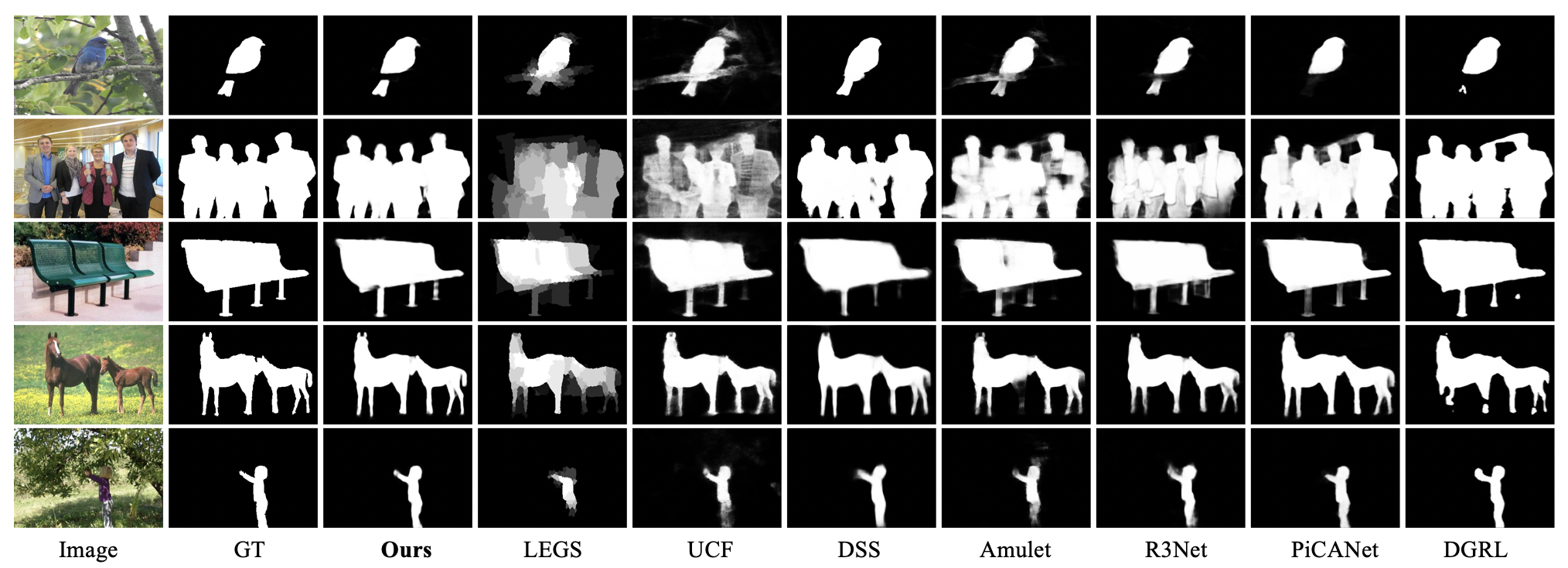}
\caption{Overall comparison of qualitative visual results between our method and selected baseline methods. It shows that our method is capable to provide more complete salient map and smooth boundaries.}
\label{fig:comparsion}
\end{figure}
\subsection{Comparisons with State-of-the-arts}
We perform our proposed method on five datasets to compare with 11 previous state-of-the-art methods, which include LEGS~\cite{wang2015deep}, UCF~\cite{zhang2017learning}, DSS~\cite{hou2017deeply}, Amulet~\cite{zhang2017amulet}, R3Net~\cite{deng2018r3net}, DGRL~\cite{wang2018detect}, PiCANet~\cite{liu2018picanet}, BMPM~\cite{zhang2018bi}, MLMSNet~\cite{wu2019mutual},
AFNet~\cite{feng2019attentive} and PAGE-Net~\cite{wang2019salient}. For fair comparisons, we use the results which are generated by their original work with default parameters and released by the authors. Moreover, all results are evaluated by the same evaluation method without any other processing tools.
\subsubsection{Quantitative Comparisons}
Figure~\ref{fig:pr_curves} and Table~\ref{tab:quantative} show the evaluation results of the proposed framework in comparison to eleven state-of-the-art methods on five challenging salient object datasets. More specifically, in Figure~\ref{fig:pr_curves}, PR curve of the proposed methodology (red line) outperforms the state-of-the-art methods. This result means that our method has better robustness than other previous methods. Furthermore, the quantitative results are listed in Table~\ref{tab:quantative}. The proposed method achieves higher F-measure scores and lower error scores than other methods, demonstrating that our novel model outperforms almost all previous state-of-the-art models on the different testing datasets.
\subsubsection{Qualitative Comparisons}
Figure~\ref{fig:comparsion} illustrates the visual comparisons in order to further show the advantages of the method. More precisely, compared to other approaches, the detection results of our method show the best performance on the different challenging scenarios. In other words, the detection results, even in certain details, are close to the ground truth.
\subsection{Ablation study}
In this subsection, we conduct a series of experiments on five different datasets to investigate the effectiveness of two modules. The ablation experiments are trained on DUTS~\cite{wang2017learning} training dataset in the same environment. From Table~\ref{tab:quantative}, the model which contains PSAM and channel-wise attention module achieves the best performance, demonstrating that the proposed modules can effectively assist the baseline's the salient object detection performance. More specifically, we initially conduct the baseline experiments on FPN baseline with ResNet-50 as backbone. This basic model can generate rough saliency map which is shown in Figure~\ref{fig:ablation}(c). Then we add pyramid self-attention module (PSAM) on the baseline and the F-measure scores increases significantly on all benchmark datasets, especially for DUTS-TE~\cite{wang2017learning} and SOD~\cite{movahedi2010design}. On this basis, we add channel-wise attention on the model to compose the proposed framework. The final best results show that the channel-wise attention modules can further increase the performance and alleviate error predictions. To this end, Figure~\ref{fig:ablation}(d) and (e) demonstrate the effectiveness of two modules respectively.
\begin{table}[h]
\begin{center}
\scalebox{0.79}{
\begin{tabular}{|c|c|c|c|c|c|c|c|c|c|c|}
\hline
\multirow{2}{*}{Method} & \multicolumn{2}{c|}{DUTS-TE} & \multicolumn{2}{c|}{ECSSD} & \multicolumn{2}{c|}{HKU-IS} & \multicolumn{2}{c|}{SOD} & \multicolumn{2}{c|}{DUT-OMRON} \\ \cline{2-11}
 & F-score & MAE & F-score & MAE & F-score & MAE & F-score & MAE & F-score & MAE\\ \hline
LEGS~\cite{wang2015deep} & 0.654 & 0.138 & 0.827 & 0.118 & 0.770 & 0.118 & 0.733 & 0.196 & 0.669 & 0.133\\
UCF~\cite{zhang2017learning} & 0.771 & 0.117 & 0.910 & 0.078 & 0.888 & 0.074 & 0.803 & 0.164 & 0.734 & 0.132\\
DSS~\cite{hou2017deeply} & 0.813  & 0.064 & 0.907 & 0.062 & 0.900 & 0.050 & 0.837 & 0.126 & 0.760 & 0.074\\
Amulet~\cite{zhang2017amulet} & 0.778 & 0.085 & 0.914 & 0.059 & 0.897 & 0.051 & 0.806 & 0.141 & 0.743 & 0.098\\
R3Net~\cite{deng2018r3net} & 0.824 & 0.066 & 0.924 & 0.056 & 0.910 & 0.047 & 0.840 & 0.136 & 0.788 & 0.071\\
PiCANet~\cite{liu2018picanet} & 0.851 & 0.054 & 0.931 & 0.046 & 0.922 & 0.042 & 0.853 & \textbf{0.102} &0.794 & 0.068\\
DGRL~\cite{wang2018detect} & 0.828 & 0.050 & 0.922 & 0.041 & 0.910 & 0.036 & 0.845 & 0.104 & 0.774 & 0.062\\
BMPM~\cite{zhang2018bi} & 0.851 & 0.048 & 0.928 & 0.045 & 0.920 & 0.039 & 0.855 & 0.107 & 0.774 & 0.064\\
PAGE-Net~\cite{wang2019salient} & 0.838 & 0.051 & 0.931 & 0.042 & 0.920 & 0.036 & 0.841 & 0.111 & 0.791 & 0.062\\
MLMSNet~\cite{wu2019mutual} & 0.852 & 0.048 & 0.928 & 0.045 & 0.920 & 0.039 & 0.855 & 0.107 & 0.774 & 0.064\\
AFNet~\cite{feng2019attentive} & 0.863 & 0.045 & 0.935 & 0.042 & 0.925 & 0.036 & 0.856 & 0.109 & 0.797 & 0.057\\ \hline\hline
\textbf{Ours} & \textbf{0.879} & \textbf{0.040} & \textbf{0.944} & \textbf{0.038} & \textbf{0.931} & \textbf{0.034} & \textbf{0.874} & 0.104 & \textbf{0.813} & \textbf{0.056}\\
Baseline & 0.856 & 0.045 & 0.933 & 0.045 & 0.921 & 0.037 & 0.848 & 0.116 & 0.785 & 0.059\\
Baseline+SA & 0.876 & 0.041 & 0.940 & 0.042 & 0.928 & 0.034 & 0.857 & 0.121 & 0.803 & 0.056\\
\hline
\end{tabular}
}
\end{center}
\caption{Quantitative results with F-score and MAE on five challenging datasets: DUTS-TE, ECSSD, HKU-IS, SOD and DUT-OMRON. Our method is compared with 11 competitive baseline methods. The last three rows of the table shows the results of the ablation studies.}
\label{tab:quantative}
\end{table}
\section{Conclusion}
In this paper, we propose a novel end-to-end salient object detection method. Considering the intrinsic problems of the FPN architecture, a pyramid self-attention module (PSAM) is designed. This module contains different self-attention layers in multiple scales, leading to capture multi-scale high-level features to make the model focus on the high-level semantic information and further enlarge the receptive field.
Furthermore, we employ the channel-wise attention in lateral connections to reduce the feature redundancy and refine prediction results. Experimental results on five challenging datasets demonstrate that our proposed model surpasses 11 state-of-the-art methods and the ablation experiments also demonstrate the effectiveness of the two modules.

\bibliographystyle{unsrt}

\begin{thebibliography}{44}
\providecommand{\natexlab}[1]{#1}
\providecommand{\url}[1]{\texttt{#1}}
\expandafter\ifx\csname urlstyle\endcsname\relax
  \providecommand{\doi}[1]{doi: #1}\else
  \providecommand{\doi}{doi: \begingroup \urlstyle{rm}\Url}\fi

\bibitem[Bello et~al.(2019)Bello, Zoph, Vaswani, Shlens, and
  Le]{bello2019attention}
I.~Bello, B.~Zoph, A.~Vaswani, J.~Shlens, and Q.~V. Le.
\newblock Attention augmented convolutional networks.
\newblock In \emph{IEEE International Conference on Computer Vision}, pages
  3286--3295, 2019.

\bibitem[Borji et~al.(2014)Borji, Cheng, Hou, Jiang, and Li]{borji2014salient}
A.~Borji, M.-M. Cheng, Q.~Hou, H.~Jiang, and J.~Li.
\newblock Salient object detection: A survey.
\newblock \emph{Computational Visual Media}, pages 1--34, 2014.

\bibitem[Chen et~al.(2017)Chen, Zhang, Xiao, Nie, Shao, Liu, and
  Chua]{chen2017sca}
L.~Chen, H.~Zhang, J.~Xiao, L.~Nie, J.~Shao, W.~Liu, and T.-S. Chua.
\newblock Sca-cnn: Spatial and channel-wise attention in convolutional networks
  for image captioning.
\newblock In \emph{IEEE Conference on Computer Vision and Pattern Recognition},
  pages 5659--5667, 2017.

\bibitem[Dalal and Triggs(2005)]{dalal2005histograms}
N.~Dalal and B.~Triggs.
\newblock Histograms of oriented gradients for human detection.
\newblock In \emph{IEEE Conference on Computer Vision and Pattern Recognition},
  volume~1, pages 886--893, 2005.

\bibitem[Deng et~al.(2018)Deng, Hu, Zhu, Xu, Qin, Han, and Heng]{deng2018r3net}
Z.~Deng, X.~Hu, L.~Zhu, X.~Xu, J.~Qin, G.~Han, and P.-A. Heng.
\newblock R3net: Recurrent residual refinement network for saliency detection.
\newblock In \emph{International Joint Conference on Artificial Intelligence},
  pages 684--690, 2018.

\bibitem[Feng et~al.(2019)Feng, Lu, and Ding]{feng2019attentive}
M.~Feng, H.~Lu, and E.~Ding.
\newblock Attentive feedback network for boundary-aware salient object
  detection.
\newblock In \emph{IEEE Conference on Computer Vision and Pattern Recognition},
  pages 1623--1632, 2019.

\bibitem[He et~al.(2016)He, Zhang, Ren, and Sun]{he2016deep}
K.~He, X.~Zhang, S.~Ren, and J.~Sun.
\newblock Deep residual learning for image recognition.
\newblock In \emph{IEEE Conference on Computer Vision and Pattern Recognition},
  pages 770--778, 2016.

\bibitem[Hong et~al.(2015)Hong, You, Kwak, and Han]{hong2015online}
S.~Hong, T.~You, S.~Kwak, and B.~Han.
\newblock Online tracking by learning discriminative saliency map with
  convolutional neural network.
\newblock In \emph{International Conference on Machine Learning}, pages
  597--606, 2015.

\bibitem[Hou et~al.(2017)Hou, Cheng, Hu, Borji, Tu, and Torr]{hou2017deeply}
Q.~Hou, M.-M. Cheng, X.~Hu, A.~Borji, Z.~Tu, and P.~H. Torr.
\newblock Deeply supervised salient object detection with short connections.
\newblock In \emph{IEEE Conference on Computer Vision and Pattern Recognition},
  pages 3203--3212, 2017.

\bibitem[Hu et~al.(2018)Hu, Shen, and Sun]{hu2018squeeze}
J.~Hu, L.~Shen, and G.~Sun.
\newblock Squeeze-and-excitation networks.
\newblock In \emph{IEEE Conference on Computer Vision and Pattern Recognition},
  pages 7132--7141, 2018.

\bibitem[Kingma and Ba(2014)]{kingma2014adam}
D.~P. Kingma and J.~Ba.
\newblock Adam: A method for stochastic optimization.
\newblock \emph{arXiv preprint arXiv:1412.6980}, 2014.

\bibitem[Krizhevsky et~al.(2012)Krizhevsky, Sutskever, and
  Hinton]{krizhevsky2012imagenet}
A.~Krizhevsky, I.~Sutskever, and G.~E. Hinton.
\newblock Imagenet classification with deep convolutional neural networks.
\newblock In \emph{Advances in Neural Information Processing Systems}, pages
  1097--1105, 2012.

\bibitem[Li and Yu(2015)]{li2015visual}
G.~Li and Y.~Yu.
\newblock Visual saliency based on multiscale deep features.
\newblock In \emph{IEEE Conference on Computer vision and Pattern Recognition},
  pages 5455--5463, 2015.

\bibitem[Lin et~al.(2017)Lin, Doll{\'a}r, Girshick, He, Hariharan, and
  Belongie]{lin2017feature}
T.-Y. Lin, P.~Doll{\'a}r, R.~Girshick, K.~He, B.~Hariharan, and S.~Belongie.
\newblock Feature pyramid networks for object detection.
\newblock In \emph{IEEE Conference on Computer Vision and Pattern Recognition},
  pages 2117--2125, 2017.

\bibitem[Liu et~al.(2019{\natexlab{a}})Liu, Hou, Cheng, Feng, and
  Jiang]{liu2019simple}
J.-J. Liu, Q.~Hou, M.-M. Cheng, J.~Feng, and J.~Jiang.
\newblock A simple pooling-based design for real-time salient object detection.
\newblock In \emph{IEEE Conference on Computer Vision and Pattern Recognition},
  pages 3917--3926, 2019{\natexlab{a}}.

\bibitem[Liu et~al.(2018)Liu, Han, and Yang]{liu2018picanet}
N.~Liu, J.~Han, and M.-H. Yang.
\newblock Picanet: Learning pixel-wise contextual attention for saliency
  detection.
\newblock In \emph{IEEE Conference on Computer Vision and Pattern Recognition},
  pages 3089--3098, 2018.

\bibitem[Liu et~al.(2019{\natexlab{b}})Liu, Huang, Dai, Ren, and
  Stathaki]{liu2019gated}
T.~Liu, J.-J. Huang, T.~Dai, G.~Ren, and T.~Stathaki.
\newblock Gated multi-layer convolutional feature extraction network for robust
  pedestrian detection.
\newblock \emph{arXiv preprint arXiv:1910.11761}, 2019{\natexlab{b}}.

\bibitem[Lowe(2004)]{lowe2004distinctive}
D.~G. Lowe.
\newblock Distinctive image features from scale-invariant keypoints.
\newblock \emph{International Journal of Computer Vision}, 60\penalty0
  (2):\penalty0 91--110, 2004.

\bibitem[Ma et~al.(2002)Ma, Lu, Zhang, and Li]{ma2002user}
Y.-F. Ma, L.~Lu, H.-J. Zhang, and M.~Li.
\newblock A user attention model for video summarization.
\newblock In \emph{International Conference on Multimedia}, pages 533--542,
  2002.

\bibitem[Movahedi and Elder(2010)]{movahedi2010design}
V.~Movahedi and J.~H. Elder.
\newblock Design and perceptual validation of performance measures for salient
  object segmentation.
\newblock In \emph{IEEE Conference on Computer Vision and Pattern Recognition},
  pages 49--56, 2010.

\bibitem[Nair and Hinton(2010)]{nair2010rectified}
V.~Nair and G.~E. Hinton.
\newblock Rectified linear units improve restricted boltzmann machines.
\newblock In \emph{International Conference on Machine Learning}, pages
  807--814, 2010.

\bibitem[Qin et~al.(2019)Qin, Zhang, Huang, Gao, Dehghan, and
  Jagersand]{qin2019basnet}
X.~Qin, Z.~Zhang, C.~Huang, C.~Gao, M.~Dehghan, and M.~Jagersand.
\newblock Basnet: Boundary-aware salient object detection.
\newblock In \emph{IEEE Conference on Computer Vision and Pattern Recognition},
  pages 7479--7489, 2019.

\bibitem[Ramachandran et~al.(2019)Ramachandran, Parmar, Vaswani, Bello,
  Levskaya, and Shlens]{ramachandran2019stand}
P.~Ramachandran, N.~Parmar, A.~Vaswani, I.~Bello, A.~Levskaya, and J.~Shlens.
\newblock Stand-alone self-attention in vision models.
\newblock \emph{arXiv preprint arXiv:1906.05909}, 2019.

\bibitem[Ren et~al.(2013)Ren, Gao, Chia, and Tsang]{ren2013region}
Z.~Ren, S.~Gao, L.-T. Chia, and I.~W.-H. Tsang.
\newblock Region-based saliency detection and its application in object
  recognition.
\newblock \emph{IEEE Transactions on Circuits and Systems for Video
  Technology}, 24\penalty0 (5):\penalty0 769--779, 2013.

\bibitem[Schillaci et~al.(2013)Schillaci, Bodiro{\v{z}}a, and
  Hafner]{schillaci2013evaluating}
G.~Schillaci, S.~Bodiro{\v{z}}a, and V.~V. Hafner.
\newblock Evaluating the effect of saliency detection and attention
  manipulation in human-robot interaction.
\newblock \emph{International Journal of Social Robotics}, 5\penalty0
  (1):\penalty0 139--152, 2013.

\bibitem[Simakov et~al.(2008)Simakov, Caspi, Shechtman, and
  Irani]{simakov2008summarizing}
D.~Simakov, Y.~Caspi, E.~Shechtman, and M.~Irani.
\newblock Summarizing visual data using bidirectional similarity.
\newblock In \emph{IEEE Conference on Computer Vision and Pattern Recognition},
  pages 1--8, 2008.

\bibitem[Vaswani et~al.(2017)Vaswani, Shazeer, Parmar, Uszkoreit, Jones, Gomez,
  Kaiser, and Polosukhin]{vaswani2017attention}
A.~Vaswani, N.~Shazeer, N.~Parmar, J.~Uszkoreit, L.~Jones, A.~N. Gomez,
  {\L}.~Kaiser, and I.~Polosukhin.
\newblock Attention is all you need.
\newblock In \emph{Advances in Neural Information Processing Systems}, pages
  5998--6008, 2017.

\bibitem[Wang et~al.(2015)Wang, Lu, Ruan, and Yang]{wang2015deep}
L.~Wang, H.~Lu, X.~Ruan, and M.-H. Yang.
\newblock Deep networks for saliency detection via local estimation and global
  search.
\newblock In \emph{IEEE Conference on Computer Vision and Pattern Recognition},
  pages 3183--3192, 2015.

\bibitem[Wang et~al.(2017)Wang, Lu, Wang, Feng, Wang, Yin, and
  Ruan]{wang2017learning}
L.~Wang, H.~Lu, Y.~Wang, M.~Feng, D.~Wang, B.~Yin, and X.~Ruan.
\newblock Learning to detect salient objects with image-level supervision.
\newblock In \emph{IEEE Conference on Computer Vision and Pattern Recognition},
  pages 136--145, 2017.

\bibitem[Wang et~al.(2018{\natexlab{a}})Wang, Zhang, Wang, Lu, Yang, Ruan, and
  Borji]{wang2018detect}
T.~Wang, L.~Zhang, S.~Wang, H.~Lu, G.~Yang, X.~Ruan, and A.~Borji.
\newblock Detect globally, refine locally: A novel approach to saliency
  detection.
\newblock In \emph{IEEE Conference on Computer Vision and Pattern Recognition},
  pages 3127--3135, 2018{\natexlab{a}}.

\bibitem[Wang et~al.(2019)Wang, Zhao, Shen, Hoi, and Borji]{wang2019salient}
W.~Wang, S.~Zhao, J.~Shen, S.~C. Hoi, and A.~Borji.
\newblock Salient object detection with pyramid attention and salient edges.
\newblock In \emph{IEEE Conference on Computer Vision and Pattern Recognition},
  pages 1448--1457, 2019.

\bibitem[Wang et~al.(2018{\natexlab{b}})Wang, You, Li, and Ma]{wang2018weakly}
X.~Wang, S.~You, X.~Li, and H.~Ma.
\newblock Weakly-supervised semantic segmentation by iteratively mining common
  object features.
\newblock In \emph{IEEE Conference on Computer Vision and Pattern Recognition},
  pages 1354--1362, 2018{\natexlab{b}}.

\bibitem[Wei et~al.(2017)Wei, Feng, Liang, Cheng, Zhao, and Yan]{wei2017object}
Y.~Wei, J.~Feng, X.~Liang, M.-M. Cheng, Y.~Zhao, and S.~Yan.
\newblock Object region mining with adversarial erasing: A simple
  classification to semantic segmentation approach.
\newblock In \emph{IEEE Conference on Computer Vision and Pattern Recognition},
  pages 1568--1576, 2017.

\bibitem[Wu et~al.(2019{\natexlab{a}})Wu, Feng, Guan, Wang, Lu, and
  Ding]{wu2019mutual}
R.~Wu, M.~Feng, W.~Guan, D.~Wang, H.~Lu, and E.~Ding.
\newblock A mutual learning method for salient object detection with
  intertwined multi-supervision.
\newblock In \emph{IEEE Conference on Computer Vision and Pattern Recognition},
  pages 8150--8159, 2019{\natexlab{a}}.

\bibitem[Wu et~al.(2019{\natexlab{b}})Wu, Su, and Huang]{wu2019cascaded}
Z.~Wu, L.~Su, and Q.~Huang.
\newblock Cascaded partial decoder for fast and accurate salient object
  detection.
\newblock In \emph{IEEE Conference on Computer Vision and Pattern Recognition},
  pages 3907--3916, 2019{\natexlab{b}}.

\bibitem[Yan et~al.(2013)Yan, Xu, Shi, and Jia]{yan2013hierarchical}
Q.~Yan, L.~Xu, J.~Shi, and J.~Jia.
\newblock Hierarchical saliency detection.
\newblock In \emph{IEEE Conference on Computer Vision and Pattern Recognition},
  pages 1155--1162, 2013.

\bibitem[Yang et~al.(2013)Yang, Zhang, Lu, Ruan, and Yang]{yang2013saliency}
C.~Yang, L.~Zhang, H.~Lu, X.~Ruan, and M.-H. Yang.
\newblock Saliency detection via graph-based manifold ranking.
\newblock In \emph{IEEE Conference on Computer Vision and Pattern Recognition},
  pages 3166--3173, 2013.

\bibitem[Yuan et~al.(2018)Yuan, Yue, and Zhang]{yuan2018rgb}
X.~Yuan, J.~Yue, and Y.~Zhang.
\newblock Rgb-d saliency detection: Dataset and algorithm for robot vision.
\newblock In \emph{International Conference on Robotics and Biomimetics}, pages
  1028--1033, 2018.

\bibitem[Zhang et~al.(2017{\natexlab{a}})Zhang, Meng, Zhao, and
  Han]{zhang2017bridging}
D.~Zhang, D.~Meng, L.~Zhao, and J.~Han.
\newblock Bridging saliency detection to weakly supervised object detection
  based on self-paced curriculum learning.
\newblock \emph{arXiv preprint arXiv:1703.01290}, 2017{\natexlab{a}}.

\bibitem[Zhang et~al.(2018)Zhang, Dai, Lu, He, and Wang]{zhang2018bi}
L.~Zhang, J.~Dai, H.~Lu, Y.~He, and G.~Wang.
\newblock A bi-directional message passing model for salient object detection.
\newblock In \emph{IEEE Conference on Computer Vision and Pattern Recognition},
  pages 1741--1750, 2018.

\bibitem[Zhang et~al.(2017{\natexlab{b}})Zhang, Wang, Lu, Wang, and
  Ruan]{zhang2017amulet}
P.~Zhang, D.~Wang, H.~Lu, H.~Wang, and X.~Ruan.
\newblock Amulet: Aggregating multi-level convolutional features for salient
  object detection.
\newblock In \emph{IEEE International Conference on Computer Vision}, pages
  202--211, 2017{\natexlab{b}}.

\bibitem[Zhang et~al.(2017{\natexlab{c}})Zhang, Wang, Lu, Wang, and
  Yin]{zhang2017learning}
P.~Zhang, D.~Wang, H.~Lu, H.~Wang, and B.~Yin.
\newblock Learning uncertain convolutional features for accurate saliency
  detection.
\newblock In \emph{IEEE International Conference on Computer Vision}, pages
  212--221, 2017{\natexlab{c}}.

\bibitem[Zhao et~al.(2015)Zhao, Ouyang, Li, and Wang]{zhao2015saliency}
R.~Zhao, W.~Ouyang, H.~Li, and X.~Wang.
\newblock Saliency detection by multi-context deep learning.
\newblock In \emph{IEEE Conference on Computer Vision and Pattern Recognition},
  pages 1265--1274, 2015.

\bibitem[Zhao and Wu(2019)]{zhao2019pyramid}
T.~Zhao and X.~Wu.
\newblock Pyramid feature attention network for saliency detection.
\newblock In \emph{IEEE Conference on Computer Vision and Pattern Recognition},
  pages 3085--3094, 2019.

\end{thebibliography}

\end{document}